\documentclass[conference]{ieeetran}
\IEEEoverridecommandlockouts

\usepackage{url}

\usepackage{makecell}

\usepackage{graphicx}

\usepackage{subcaption}
\usepackage[utf8]{inputenc}
\usepackage[T1]{fontenc}

\usepackage{textcomp}

\usepackage[plainpages=false,hypertexnames=false,pdfpagelabels]{hyperref} 

\usepackage[dvipsnames]{xcolor}

\begin{document}

\title{The utilization of spherical camera in simulation for service robotics
\thanks{This work was funded within the INCARE AAL-2017-059 project ,,Integrated Solution for Innovative Elderly Care'' by the AAL JP and co-funded by the AAL JP countries (National Centre for Research and Development, Poland under Grant AAL2/2/INCARE/2018).}}

\author{\IEEEauthorblockN{Krystian Chachuła}
		\IEEEauthorblockA{
			Warsaw University of Technology\\
			Institute of Control and Computation Engineering\\
			Warsaw, Poland,\\
			Email: krystian.chachula.stud@pw.edu.pl}
		\and
		\IEEEauthorblockN{Maciej Stefańczyk}
		\IEEEauthorblockA{
			Warsaw University of Technology\\
			Institute of Control and Computation Engineering\\
            Warsaw, Poland,\\
			Email: maciej.stefanczyk@pw.edu.pl}
}



\maketitle

\begin{abstract}

    Safety is one of the most critical factors in robotics, especially when robots have to 
    collaborate with people in a~shared environment. Testing the physical systems, however, must
    focus on much more than just software. One of the common steps in robotic system development
    is the utilization of simulators, which are very good for tasks like navigation or manipulation.
    Testing vision systems is more challenging, as the simulated data often is far from the
    real camera readings. In this paper, we show the advantages of using the spherical camera
    for recording the sequences of test images and a~way to integrate those with existing
    robotic simulator. The presented system also has the possibility to be extended with rendered
    objects to further improve its usability.



\end{abstract}

\section{Introduction}

\subsection{The problem}

People use their sight as the most important sense when they move through their surroundings.
It is therefore essential for service robots to use image processing.
When robots coexist with humans, safety is highly important \cite{haddadin2007safety}.
One of many measures to be taken is software testing,
which requires test data and, in the domain of computer vision, this data consists of images or videos from a~robot's camera \cite{7759425}.
However, using a~real robot for testing can be dangerous.
The danger stems from the fact that, while testing, the robot is controlled by yet untested software, which, in case of service robots, can be complicated \cite{Dudek-multitasking-romoco-2019-twiki}. This can lead to damaging the environment, the robot itself or in the worst case scenario can harm the people.
Using a~real robot for testing can also be tedious -- after each test iteration one has to return the environment to the initial state and move robot back to the staring pose.
To speed up the process it is possible to run the application on multiple robots \cite{levine2018learning}, but this is not the cost-effective solution.
This creates the need to simulate the tested robot~\cite{seredynski2016graph} and, specifically, its one or more cameras. 
In this article, we examine the existing robotic simulators as a~tool for testing vision algorithms and propose
their extension with the use of spherical images.

\subsection{State of the art}

\begin{figure}[!ht]
    \includegraphics[width=0.494\columnwidth]{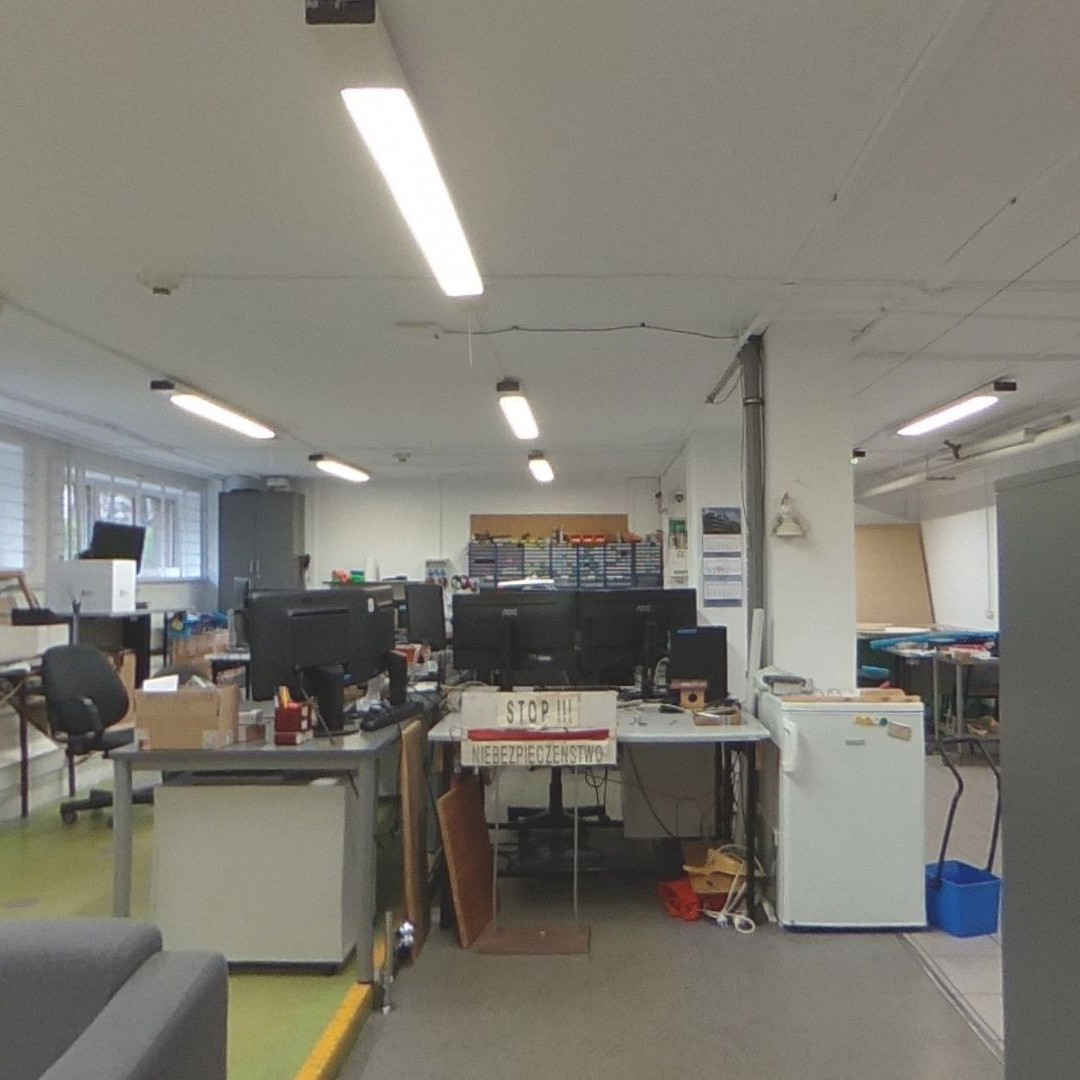}\hfill%
    \includegraphics[width=0.494\columnwidth]{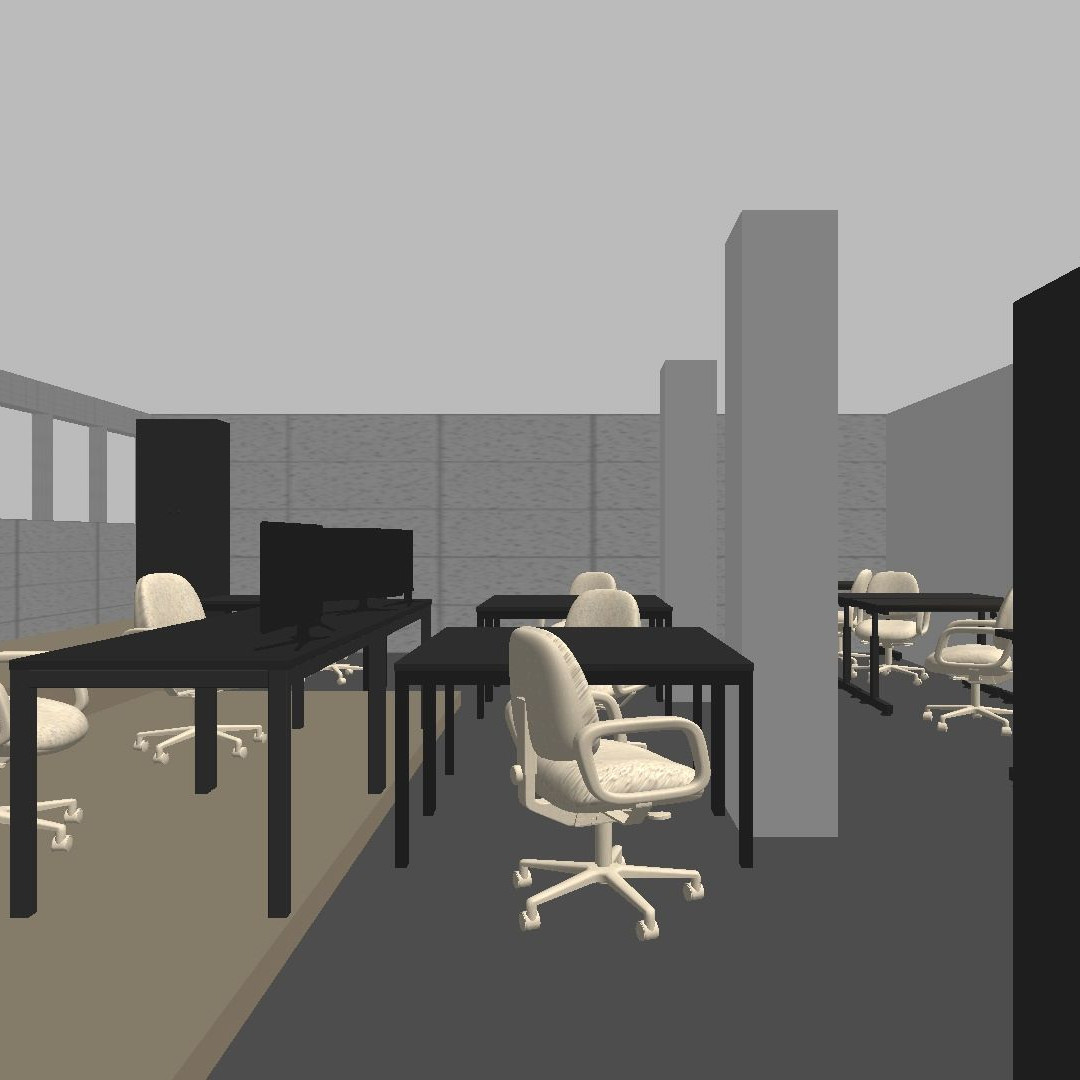}
    \caption{A~real (left) and a~simulated (right) camera in Gazebo}
    \label{fig:gazebo_vs_real}
\end{figure}

A~practical method of gathering image data for testing is recording a~video while the robot is executing a~given path in the testing environment for future use.
Another way is using simulators such as Gazebo \cite{koenig2004design} or UnrealROX \cite{martinez2019unrealrox}.
Gazebo is meant to be a~universal simulator, with physics and 3D rendering engine. It allows for accurate 
robots and environment modeling and robot behavior simulation. Its integration with the Robot Operating \
System makes it very popular amongst the research and industry community. It lacks the most in the vision 
field. Although the 3D rendering is possible, environments created in the simulation are often filled
with repetitive objects with very simple textures (Fig.~\ref{fig:gazebo_vs_real}).
 The lighting system is also very simplified, without any 
shadows and reflections, which makes the renders unsuitable for the vision algorithms evaluation. 

\begin{figure}[!htb]
    \includegraphics[width=0.494\columnwidth]{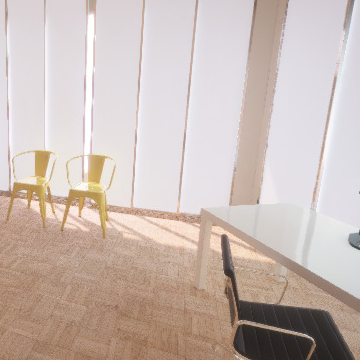}\hfill%
    \includegraphics[width=0.494\columnwidth]{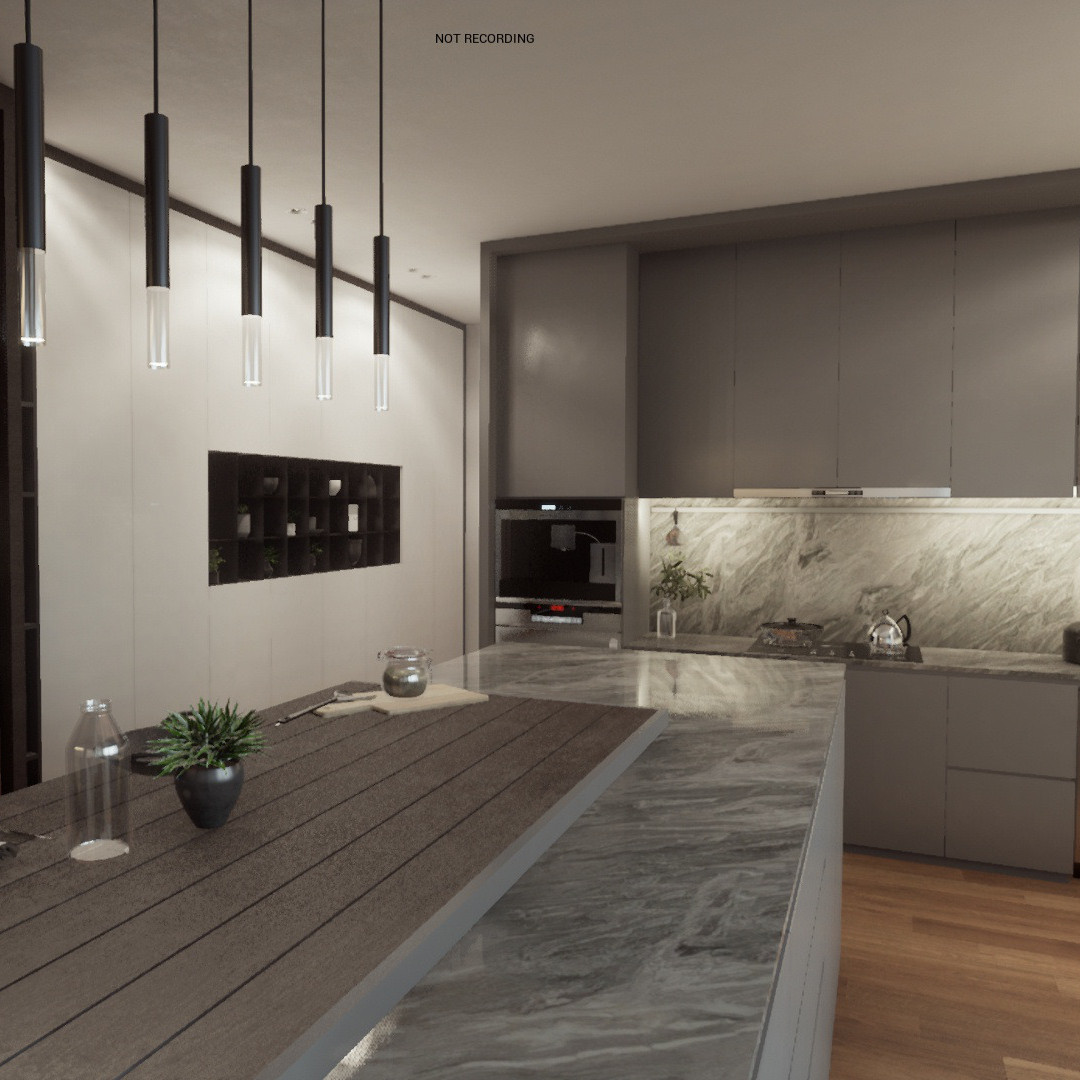}
    \caption{Images generated by the UnrealROX simulator \cite{martinez2019unrealrox}}
    \label{fig:rox_samples}
\end{figure}

On the other hand, simulators like UnrealROX are designed to be used as a~tool for training visual algorithms,
especially those based on deep learning, requiring a~lot of training data. Simulator, based on Unreal Engine, 
is able to produce photo-realistic scenes in real-time (Fig.~\ref{fig:rox_samples}). 
To prepare such scenes, 
however, an artist is required most of the time. A~solution like this would be perfect, but lack of integration 
(and no possibility thereof) with robotics systems makes it a~tool only for off-line learning, not interactive
testing.

From this short review, we can point out some features that are desired in simulators \cite{staranowicz2011survey} 
apart from the photorealism.
One is the capability to modify the world model used for image generation quickly.
This allows the tester to create new test scenarios rapidly.
Another essential feature is the ability to edit the robot's path dynamically.
Of course, this is not possible when simulating using a~prerecorded video, but should be achievable 
in every generic simulator, as well as the capability of simulating robot's interactions with the
environment, such as collisions and manipulation. Integration with ROS is also important. From the summary
in Tab.~\ref{tab:simulation_methods} we can see that fulfilling all the requirements at once is hard to achieve.

\begin{table}[!ht]
    \caption{Comparison of different camera simulation methods}
    \label{tab:simulation_methods}
    \centering
    \setlength{\tabcolsep}{1em}
    \def\arraystretch{1.2}
    \begin{tabular}{ |c|c|c|c|c| } 
        \hline
        Feature & \textit{Gazebo} & \textit{video} & \textit{UnrealROX} \\ 
        \hline
        \makecell{environment modification} & yes & \textbf{no} & yes \\
        \makecell{editable robot path} & yes & \textbf{no} & yes \\
        \makecell{interaction with the environment} & yes & \textbf{no} & yes \\
        photorealism & \textbf{no} & yes & yes \\
        ROS integration & yes & yes & \textbf{no} \\
        \hline
    \end{tabular}
\end{table}

\subsection{The solution}

\begin{figure}[ht!]
\includegraphics{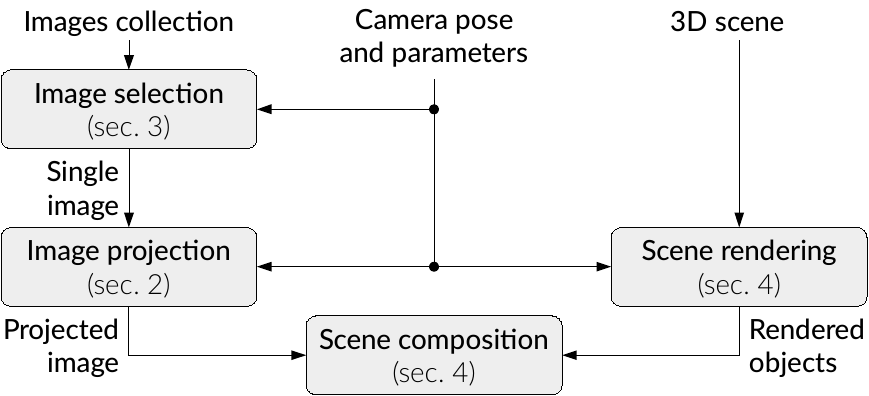}
\caption{Proposed simulation system structure}
\label{fig:flow}
\end{figure}

Using pre-recorded video/images has the biggest photorealism (as those are photos). Using the 3D simulators gives 
possibility of free robot movement and interaction with the environment. To combine both worlds,
we study the possibility of using the photos as a~source for the camera image in the simulation. 
To give the robot the possibility to look around, photos should cover the full sphere (360\textdegree\ panorama or 
full spherical photo). To have freedom of movement images from (almost) all possible robot locations should 
be available. The position of the virtual camera can be used to select exactly one (the closest) spherical image from the set.
Finally, to not limit the solution to only a~single pre-recorded environment state, there has to be
a~way of adding the simulated/rendered objects to the scenes. This will also allow for interaction. Schematic
diagram of the proposed system is presented in Fig.~\ref{fig:flow}.

The rest of the paper is structured as follows. In section~\ref{sec:spherical}, we show the methods of creating
a~single photo from a~spherical image. Section~\ref{sec:iterpolate} introduces a~way of dealing with the robot movement.
Integration with the 3D simulator and merging with the rendered overlays are presented in section~\ref{sec:gazebo}.
A~case study for a~prepared system is presented in section~\ref{sec:evaluation}, followed by the conclusions.




There are different camera types used throughout the article. A~\textbf{spherical camera} is a~type of camera with a~field of view that covers the entire sphere, and a~\textbf{regular camera} is a~type of camera with a~field of view close to $90$ degrees.
The camera that belongs to a~robot in simulation is referred to as a~\textbf{virtual camera}.
A~\textbf{projection camera} is an entity that projects a~3D scene to a~2D image in computer graphics.

\section{Transforming spherical images}
\label{sec:spherical}

\begin{figure}[ht!]
    \centering
    \includegraphics[width=0.5\columnwidth]{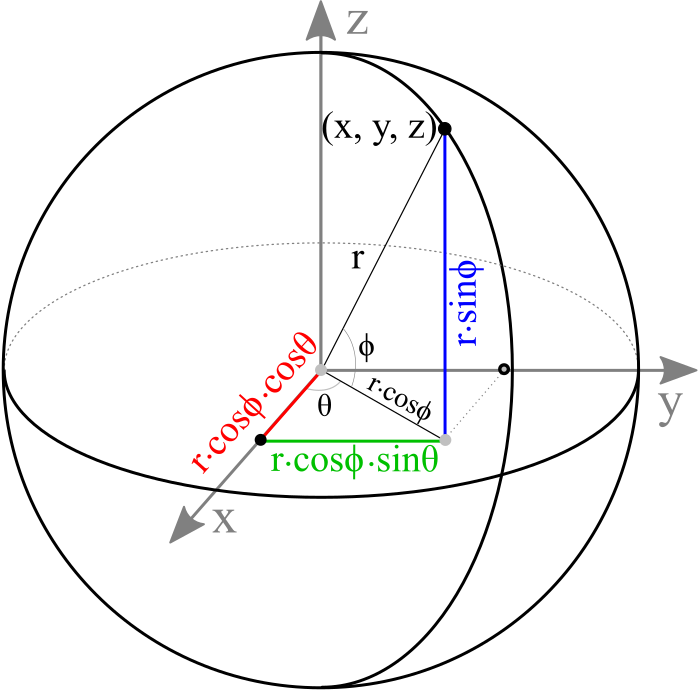}
    \caption{Sphere coordinates generation}
    \label{fig:spheresa}
\end{figure}

In order to transform a~spherical image into a~regular image, one can utilize 3D computer graphics.
Not only is this method simple, but its implementation can utilize hardware acceleration.
A~sphere is a~3D closed surface where every point on the sphere is the same distance from its origin. 
Since only a~finite number of points can be drawn, a~sphere will need to be approximated by drawing a~subset of its points.
Those points can be selected by creating a~UV Sphere.
A~UV Sphere consists of vertices that are created by performing angular steps in the spherical coordinate system (Fig.~\ref{fig:spheresa}).
Cartesian points are generated using these equations:
\begin{equation}
    x = r \cdot \cos \phi \cdot \cos \theta \quad y = r \cdot \cos \phi \cdot \sin \theta \quad z~= r \cdot \sin \phi \nonumber
\end{equation}
where $r$ is the radius of the sphere, $\phi \in [0, \pi]$ -- polar angle and $\theta \in [0, 2\pi)$ -- azimuthal angle.


\begin{figure}[ht!]
    \centering
    a) \hspace{-3mm}\includegraphics[width=0.48\columnwidth]{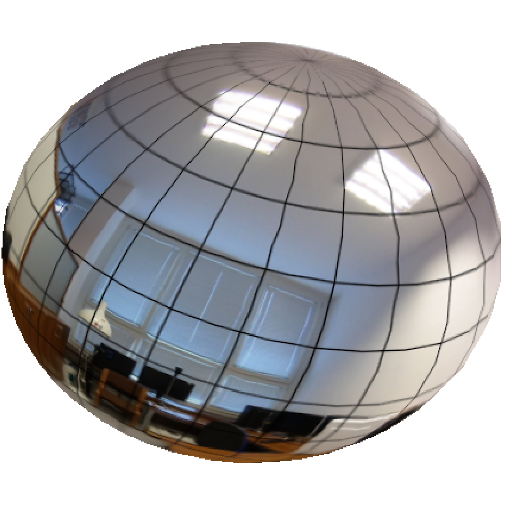}
    b) \hspace{-3mm}\includegraphics[width=0.48\columnwidth]{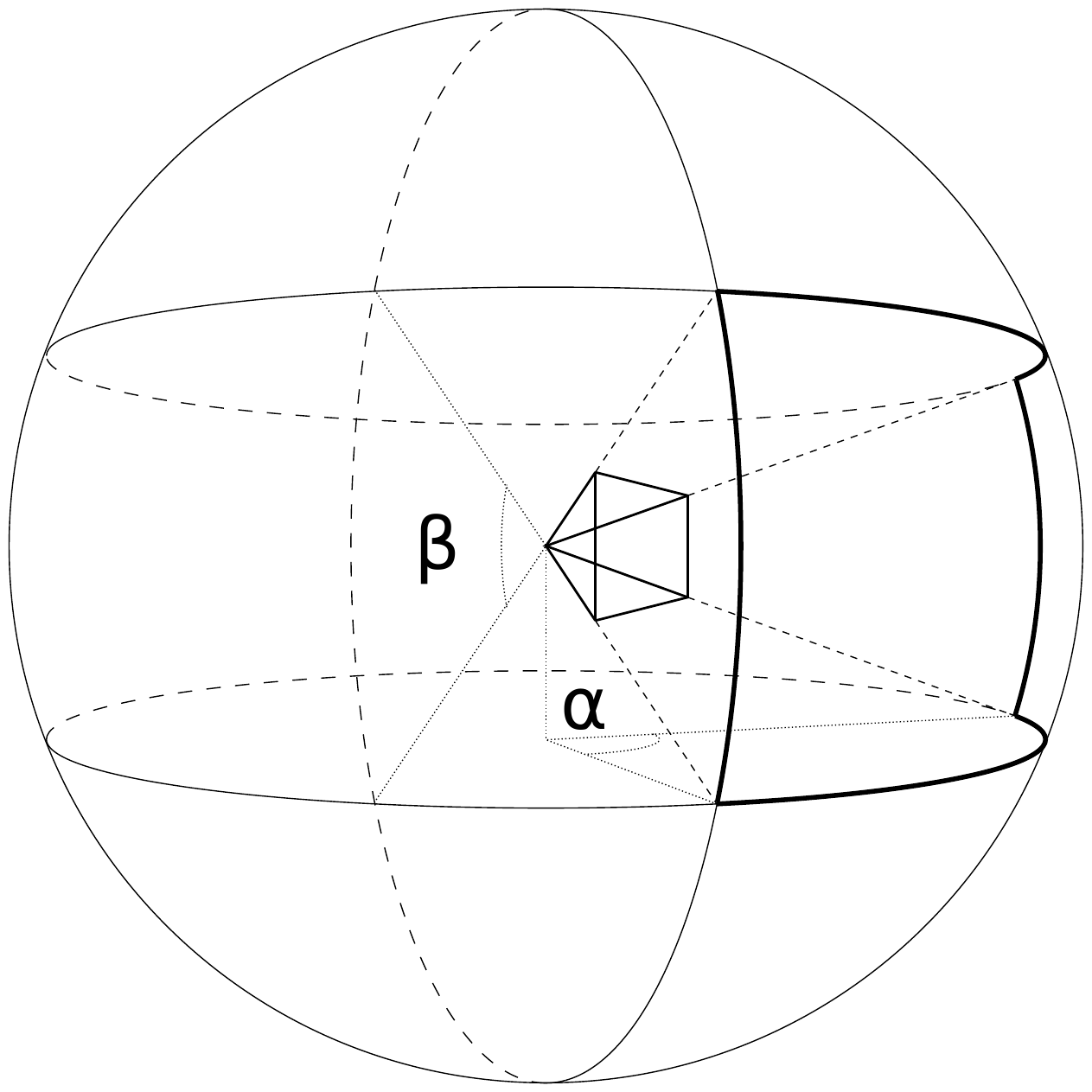}
    \caption{Sphere textured with equirectangular photo}
    \label{fig:spheresb}
\end{figure}

Then the sphere has to be textured with the spherical image (Fig.~\ref{fig:spheresb}a).
Most spherical cameras create images in the form of equirectangular projections, and, without the loss
of the generality of the solution, we assumed images in that format are available.
The process of texturing a~UV sphere with an image in the form of this projection is fairly straightforward \cite{greene1986environment}. 
Every point on the sphere can be mapped to a~point on the texture by simply assigning angles to texture 
coordinates $x_t = \theta$ and $y_t = \phi$.

Having textured the sphere, now a~projection camera can be placed in its center (Fig.~\ref{fig:spheresb}b).
Its parameters (field of view and resolution) are set to match the robot's virtual camera.
The image from this camera is a~regular image as seen by the robot looking in a~given direction.

\section{Interpolation}
\label{sec:iterpolate}

The fact that always the nearest spherical image is used for rendering poses a~significant issue.
If two virtual camera positions are close to each other it is possible, that both of them will use
the same spherical image to create the projected camera view.
Moreover, if those cameras have the same orientation relative to the global reference frame then
exactly the same regular image would be produced for both cameras.

\begin{figure}[!ht]
    \centering
    \includegraphics{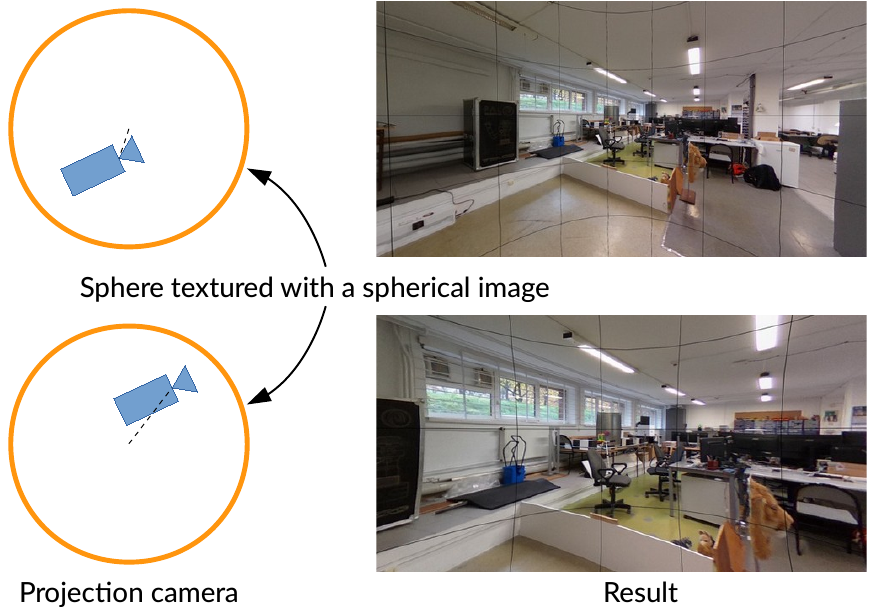}
    \caption{Tthe method of interpolating camera views inside the single spherical image}
    \label{fig:interpolation}
\end{figure}

A~simple solution to this problem would be to mimic the position of the virtual camera relative 
to a~region with the position of the projection camera, thus creating the illusion of moving closer 
to certain objects as visualized in Fig.~\ref{fig:interpolation}. This solution is not perfect,
as there will be no parallax visible, but with the photos sampled sufficiently dense the effect is satisfactory.

\section{3D objects overlay}
\label{sec:gazebo}

\begin{figure}[!ht]
    \centering
        \includegraphics[width=\columnwidth]{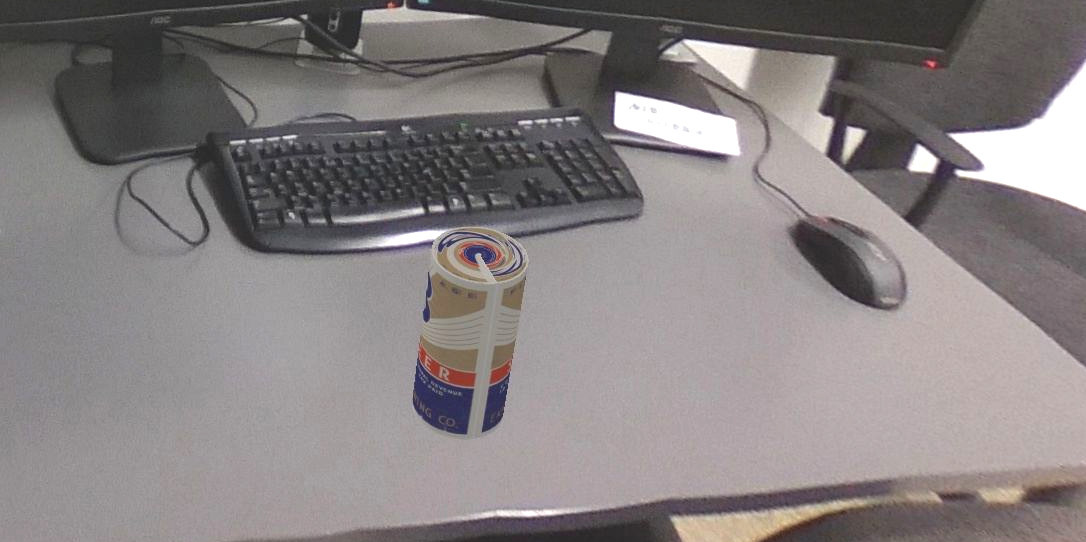}
    \caption{Rendered object (can) overelaid on the photo}
    \label{fig:join}
    \vspace{-5mm}
\end{figure}

So far, the robot can look around and move through the prepared environment. 
Generated images can be further processed in order to make this solution more versatile.
As the position and orientation of virtual camera is known, and its parameters are the same
as the projection camera, objects generated using 3D rendering software can be overlaid onto the photo.
Fig. \ref{fig:join} depicts an example of such a~process.

In our solution, we decided to stick with the Gazebo simulator for this purpose.
In order to remove unwanted objects from the image, additional plugin to the Gazebo simulator
was prepared to hide the objects from rendering in a~virtual camera, keeping them 
visible for other sensors (like laser scanner or sonars), thus keeping the obstacle avoidance
still working. This provides additional simulation features mentioned in the introduction. 
It enables the tester to modify the testing environment and provides the means of interaction 
of the simulated robot with the environment utilizing the Gazebo simulator.

\section{Evaluation}
\label{sec:evaluation}

The simulation system presented above was tested in our robotics laboratory. At first, individual modules
and system parameters were evaluated, followed by the partial scan of the room. Prepared environment is 
meant to be used with the TIAGo robot applied to service tasks related to Elderly care and assistance.

\begin{figure*}[!ht]
    \centering
    \includegraphics[width=0.33\textwidth]{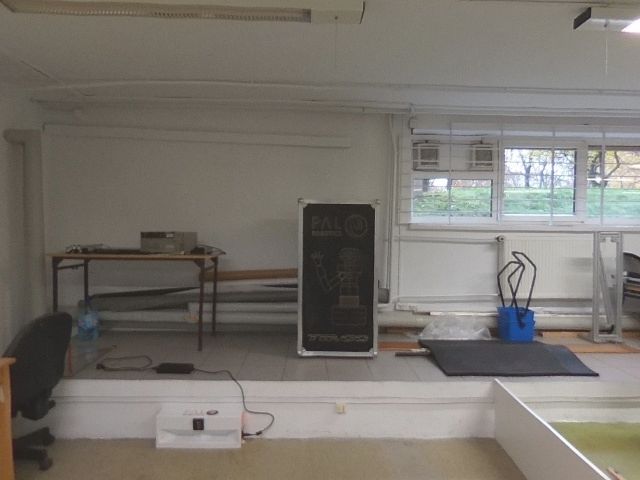}\hfill%
    \includegraphics[width=0.33\textwidth]{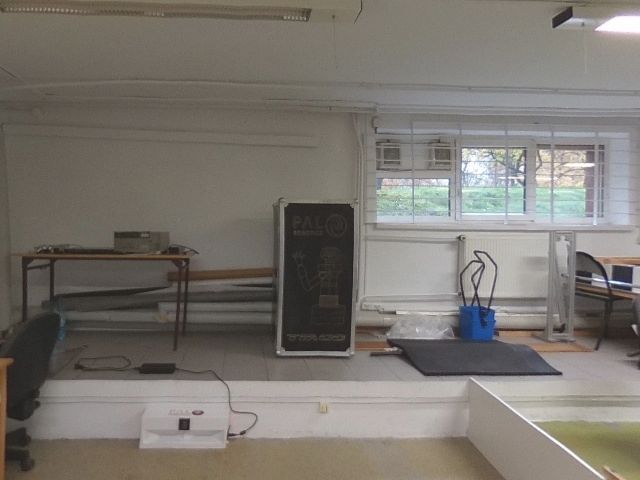}\hfill%
    \includegraphics[width=0.33\textwidth]{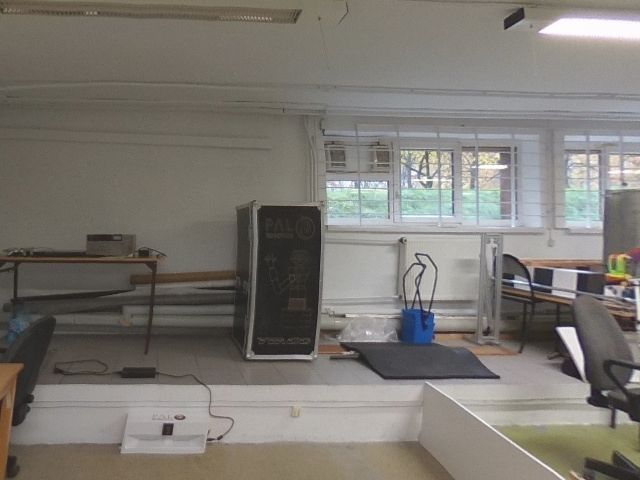}\\
    \caption{Example images generated with grid resolutions of 20, 100 and 200 cm.}
    \label{fig:grid_sizes}
\end{figure*}

\subsection{Camera selection}

In order to acquire a~full spherical photo, a~dedicated camera setup is required. Nowadays, there are different 
possibilities to achieve that, without the need for building multiple camera rig from scratch. If the price 
doesn't matter, Ladybug5+ from Flir is the option. Its precalibrated six cameras provide 360\textdegree\ 
panorama with 8K resolution. Custom solutions use single camera with specially shaped mirror \cite{wkasik2017embedded}
to achieve the omnidirectional field of view. On the consumer side of the market, there are a~few different devices to choose from,
most of them built from two fish-eye cameras facing in opposite directions. We chose the Theta V camera from Ricoh. 
This camera is capable of on-board image stitching to the equirectangular projection with final resolution
$5376\times2688$ pixels (created from two 4K photos). Another useful feature is the possibility to capture
and transfer photos using WiFi. The average angular resolution of the selected camera is around 15 px/\textdegree, 
which is more than the 10 px/\textdegree\ of a~robot's camera. 

\subsection{Rendering performance}

All the advantages of even the most advanced and photorealistic simulation are useless if its performance 
doesn't allow for (near)real-time work. There are two independent factors in photo rendering that can limit
the simulation performance. When the robot moves inside a~single cell (in range of a~single spherical
photo), only the rendering time is important. For our system, it takes on average 0.365 ms to
render a~single VGA view. This allows us to simulate the camera with 2740 FPS. 

The other latency is a~result of loading the spherical photo from disk, decoding it, and loading it into the memory of the video card.
A~single 5376 $\times$ 2688 pixels JPEG image takes on average $t = 198.672\,ms$ to load and decode on our system.
This further limits the FPS to 5 if the images are loaded synchronously.
However, if the images can be loaded in advance, this would mitigate the negative impact of the load latency on the robot's maximum speed.
In this case simulated robot's maximum speed $v_{max}$ depends on the amount of primary memory.

To achieve scaling $v_{max}$ with memory, one can utilize a~pool of threads or processes.
Now a~significant problem is choosing which spherical images should be cached in RAM.
The most straightforward strategy is to always store four images that were taken closest to the current position of the robot.
Such a~solution often loads many images that will not be used for simulation.
A~more sophisticated method would be to predict the movement of the robot.
For example, the robot's velocity vector could be projected forward until it intersects the border of a~cell.
Then, this image would be decoded and loaded in advance.

With just one thread or process for decoding images, the grid resolution of $20\,cm$ and image load time of $200\,ms$, the theoretical aggregated speed limit of the simulated robot is equal to $1\,\frac{m}{s}$.
According to TIAGo's datasheet, this value coincides with real robot's maximum speed.

\subsection{Grid resolution and interpolation parameters}

Proper distance between consecutive photos is not easy to decide and depends on the
actual environment. Photos captured to sparse will result in big jumps when the camera
moves between different grid points. Too dense and image loading and decoding
times will limit the speed of the robot in simulation to unacceptable values.

In order to make such a~decision, we created a~relatively small but dense spherical photo set.
After that, example images were generated for a~given position and orientation of the virtual camera using that set.
Next, we made the grid more sparse by selecting every second, third, and so forth photo from the set and generated images for each step.
Some of the generated images can be seen in Fig. \ref{fig:grid_sizes} (notice the vertical line distortions for too large step).
The grid resolution of 20 cm provided a~good balance between the quality of the simulation and the time needed for the creation of the photo set.

\subsection{Collecting the photos of the target environment}

\begin{figure}[!ht]
    \centering
    \includegraphics[width=\columnwidth]{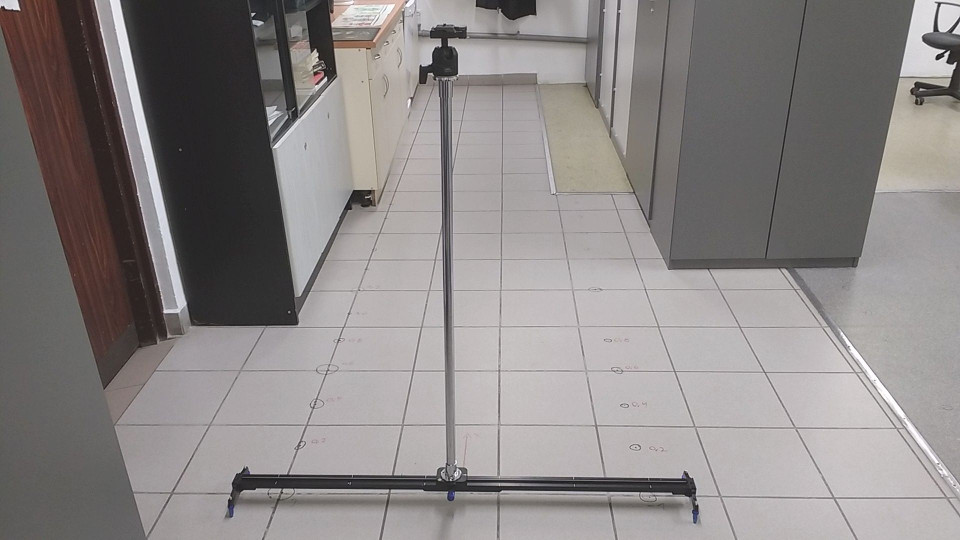}\\
    \includegraphics[width=\columnwidth]{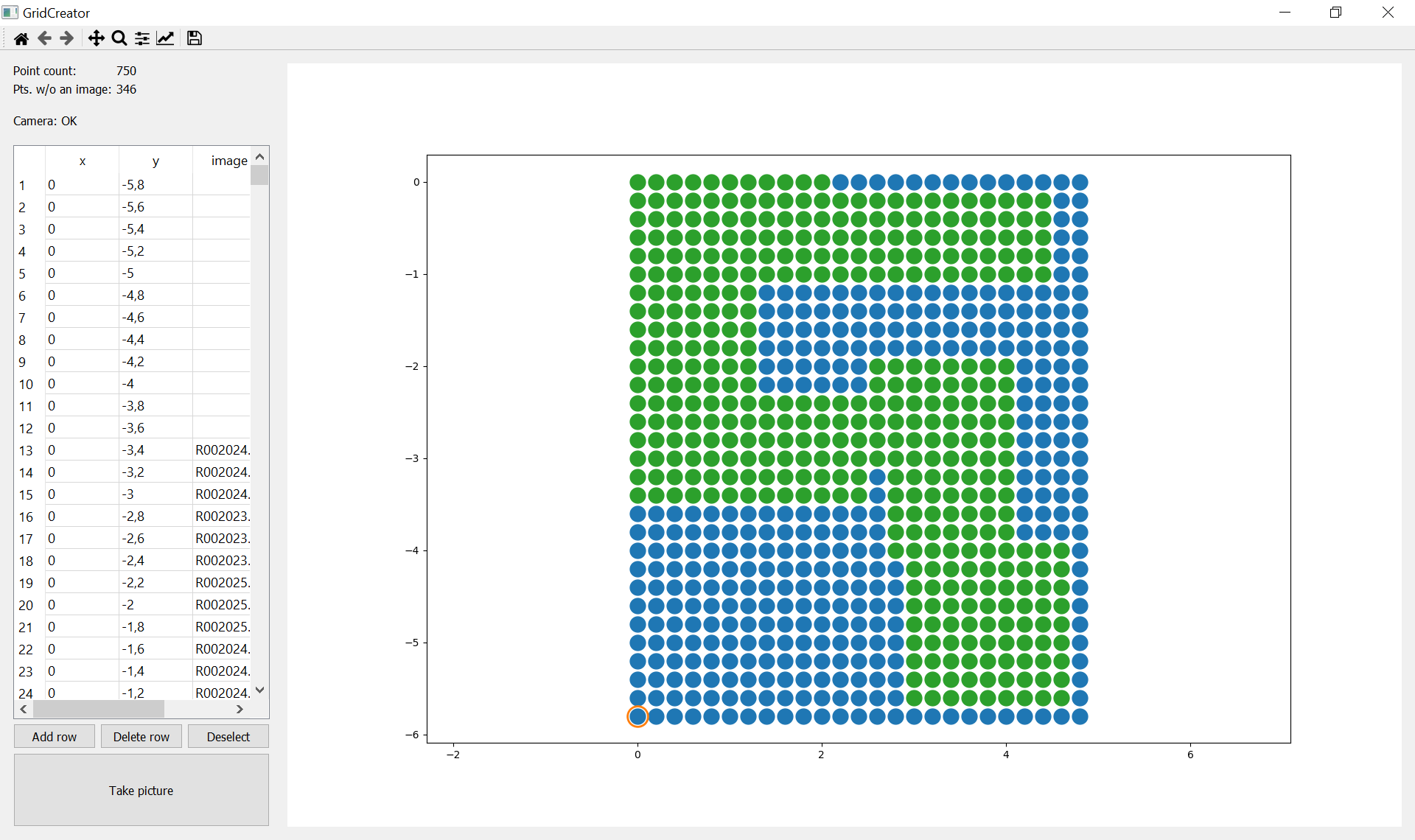}\\
    \caption{Hardware and software used for the creation of the image set}
    \label{fig:rig}
\end{figure}

Finally, when all the parameters are selected, a~set of spherical images can be created.
To facilitate the whole process, we've created helper hardware and software.
For accurate photo capture and straight camera movement, the device in Fig. \ref{fig:rig} was used.
It consists of a~rail and a~cart to which a~monopod has been fixed.
At the other end of the monopod, there is a~mounting point for the camera.
A~grid marks were put on the rail in order to allow precise positioning of the cart.
Similar marks were put on the floor in order to guide the rail.

To simplify the process and keep track of photos that are already taken and still waiting to 
be captured, an application was created (Fig. \ref{fig:rig}).
It allows for selecting points and sending requests to the camera for taking pictures using the
Open Spherical Camera API. It also automatically tags the images with the correct position.
In the selected room, 404 pictures were taken in $20\,cm$ intervals, covering the living room part,
the kitchen, and part of the corridor. The whole process took 2 hours and 17 minutes,
which yields approximately 3 pictures taken per minute. 

\begin{figure*}[!p]
    \centering
    \includegraphics[width=\textwidth]{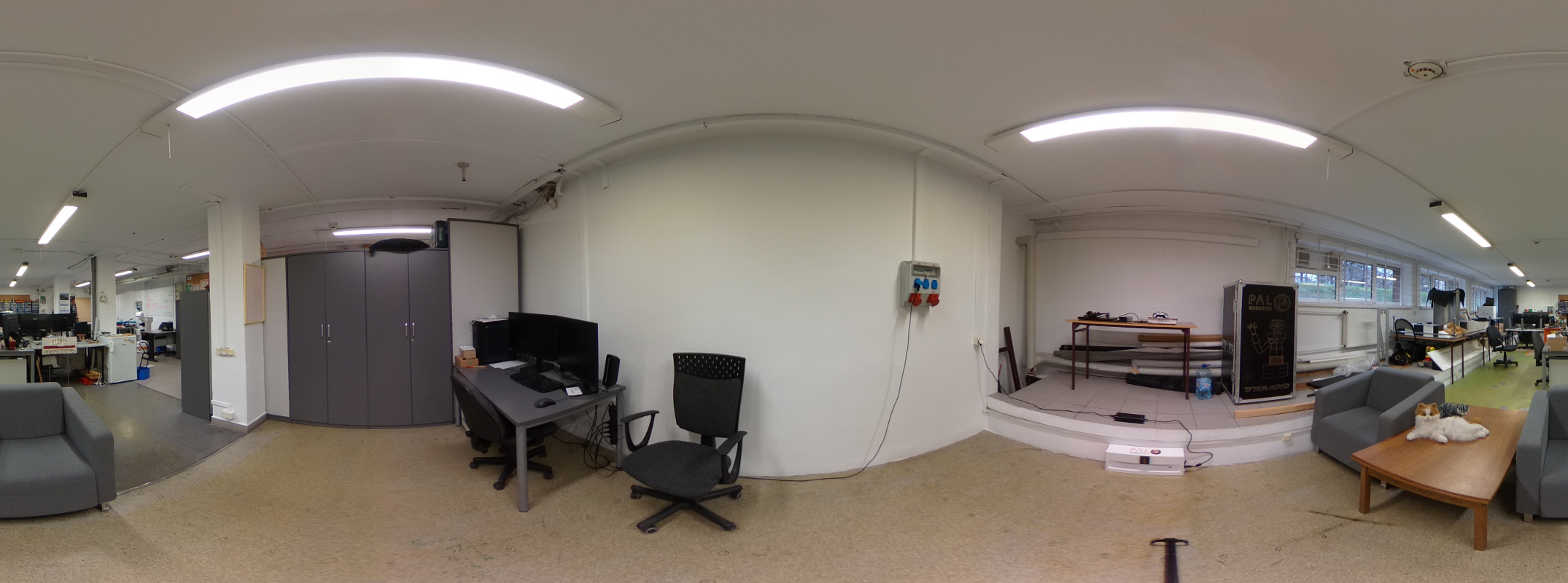}\\\vspace{1mm}
    \includegraphics[width=\textwidth]{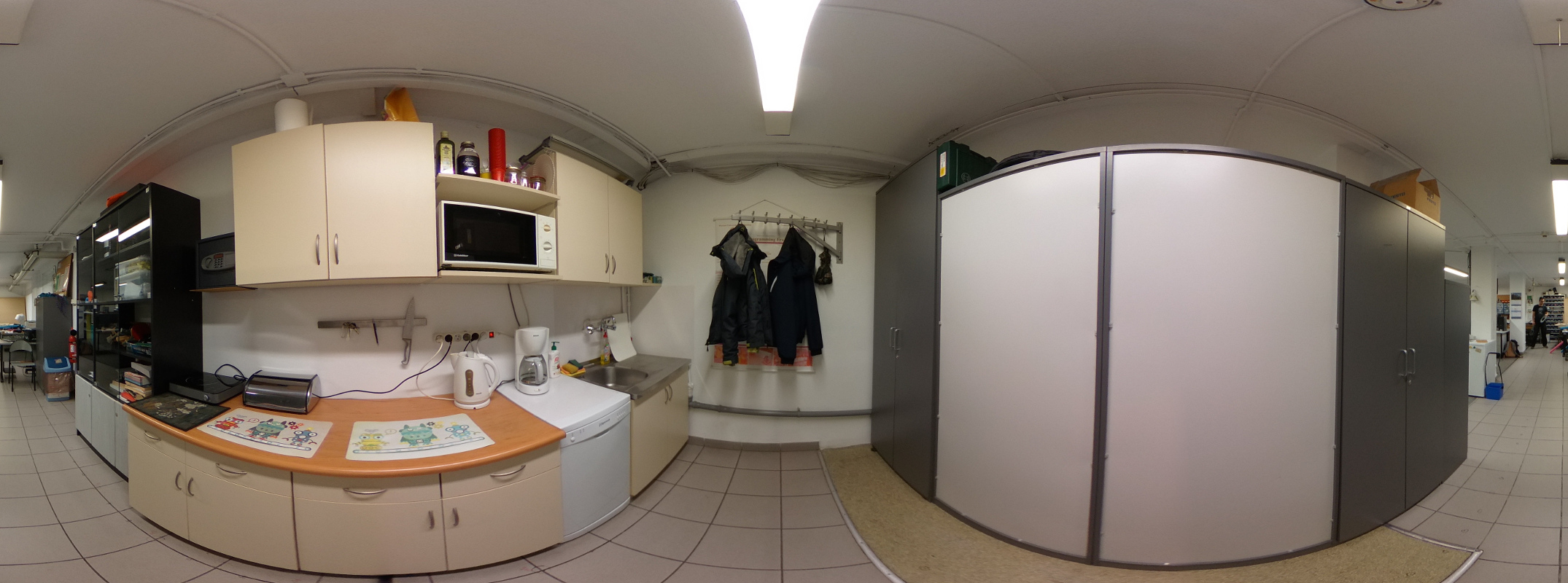}\\\vspace{1mm}
    \includegraphics[width=0.33\textwidth]{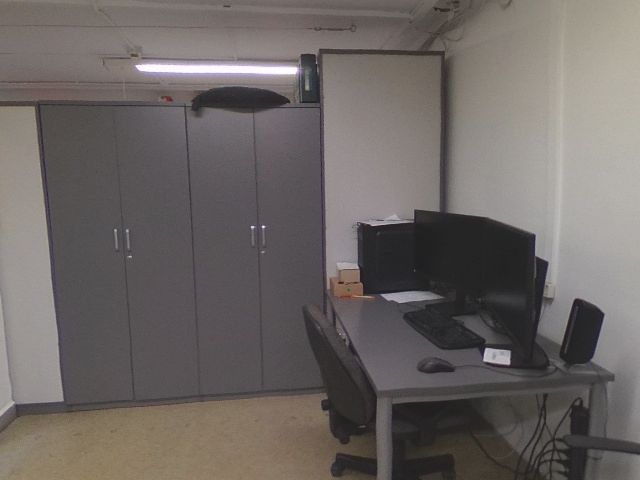}\hfill%
    \includegraphics[width=0.33\textwidth]{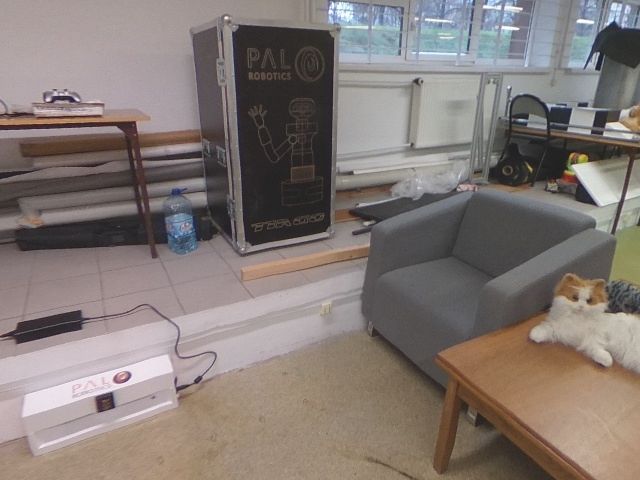}\hfill%
    \includegraphics[width=0.33\textwidth]{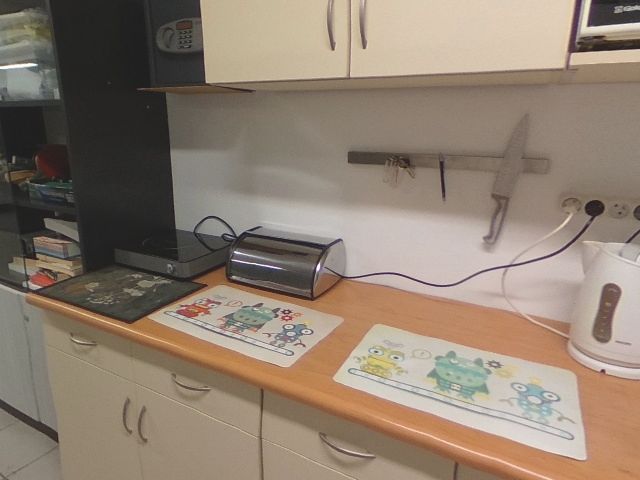}\\\vspace{1mm}
    \includegraphics[width=0.33\textwidth]{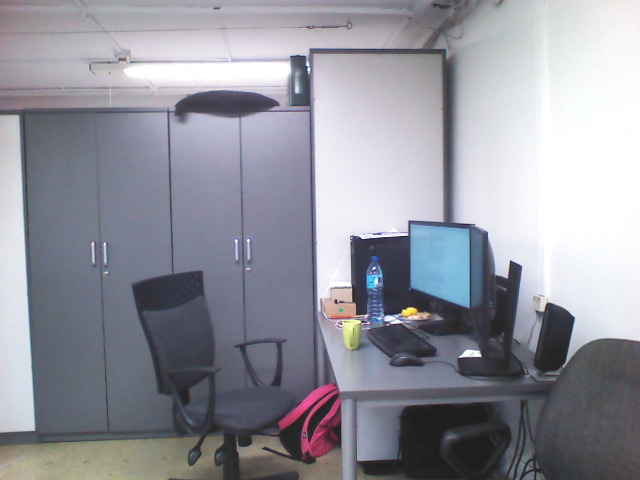}\hfill%
    \includegraphics[width=0.33\textwidth]{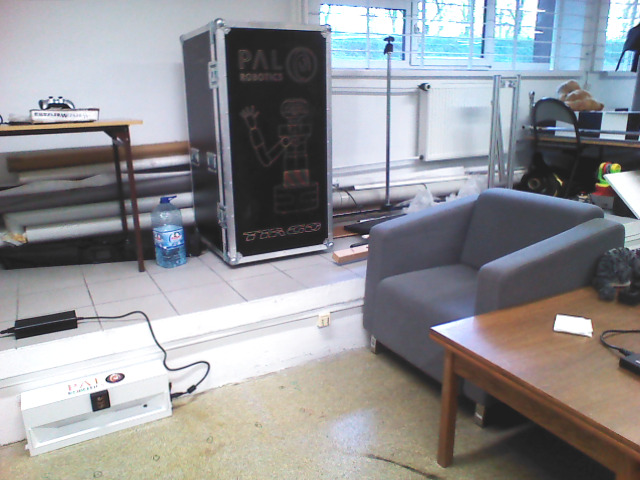}\hfill%
    \includegraphics[width=0.33\textwidth]{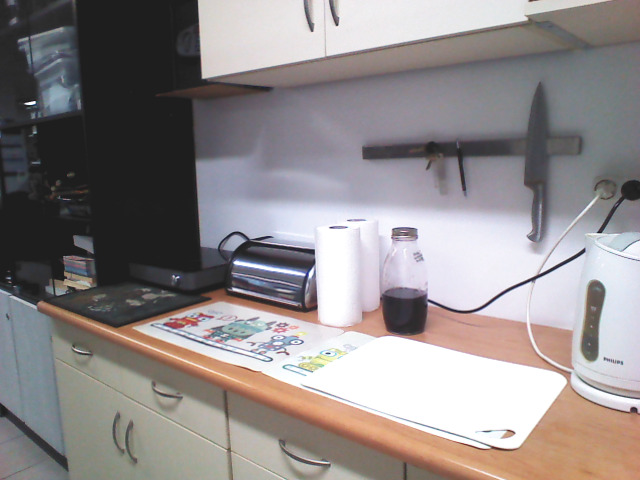}\\
    \caption{Images generated in simulation (first row) and taken by the robot (second row). Last two rows presents
    sample source spherical photos.}
    \label{fig:sim_vs_tiago}
\end{figure*}

The whole solution was evaluated by comparing generated images with images from a~real robot.
For this purpose, the TIAGo robot \cite{pages2016tiago} was used.
It was booted up and manually controlled in order to take pictures.
Then the proposed solution was used to generate images corresponding to pictures taken by the TIAGo.
Finally, images from both sources were compared, as seen in Fig. \ref{fig:sim_vs_tiago}. The images 
differ mainly in color appearance and are otherwise nearly identical. The mentioned color issue can 
be resolved by the means of color correction.

\section{Conclusions}

The article presented the idea and application for the spherical camera-based 
simulator for service robots. The system was tested by comparing images generated in simulation
with those gathered by the real robot. Those results prove that using the proposed solution a~set 
of spherical images can be used to replace a~real robot in software testing.
One important feature of presented system is that it doesn't replace existing simulators,
but rather coexists with them and provides additional features.
Presented system in its current form can be utilized to test the robotic vision algorithms
involving visual inspection \cite{mmar_winiarski_automated-2016} or even objects
grasping \cite{seredynski2016fast} when the 3D overlays are used. 

There are some issues in the system that can be improved. The process of photo acquisition
at the moment is highly manual labor. For each photo, one has to move the camera, hide from
the camera view, and capture a~photo. In its current form, it takes around 8 minutes per square meter.
The natural extension would be to utilize a~robotic platform to move the camera around and
take photos automatically. 

Manual photo capturing results also in slight misalignments between consecutive frames.
Photo stabilization techniques could be implemented to calculate angular offsets for each
image.

Created photo collection could be extended with depth information calculated from multiview
structure estimation. This would make the interpolation process work better and allow for, 
possibly, grid with bigger steps.


\bibliographystyle{IEEEtran} 
\bibliography{bibliography}

\end{document}